\documentclass{bmvc2k}

\usepackage{xcolor}
\usepackage{customization}
\usepackage{multirow}
\usepackage{graphicx}
\usepackage{makecell}
\usepackage{wrapfig}

\title{Towards Open-Vocabulary Multimodal 3D Object Detection with Attributes}

\addauthor{Xinhao Xiang$^\star$}{xhxiang@ucdavis.edu}{1}
\addauthor{Kuan-Chuan Peng}{kpeng@merl.com}{2}
\addauthor{Suhas Lohit}{slohit@merl.com}{2}
\addauthor{Michael J. Jones}{mjones@merl.com}{2}
\addauthor{Jiawei Zhang}{jiwzhang@ucdavis.edu}{1}

\addinstitution{
 University of California, Davis\\
 Davis, CA, USA
}
\addinstitution{
 Mitsubishi Electric Research Laboratories (MERL)\\
 Cambridge, MA, USA
}

\runninghead{Xiang et al.}{Open-Vocabulary Object Detection with Attributes}

\def\eg{\emph{e.g}\bmvaOneDot}

\newcommand\blfootnote[1]{
  \begin{NoHyper}
  \begingroup
  \renewcommand\thefootnote{}\footnote{#1}
  \addtocounter{footnote}{-1}
  \endgroup
  \end{NoHyper}
}

\begin{document}

\maketitle

\begin{abstract}
3D object detection plays a crucial role in autonomous systems, yet existing methods are limited by closed-set assumptions and struggle to recognize novel objects and their attributes in real-world scenarios. 
We propose \ours, a novel framework enabling both open-vocabulary 3D object and attribute detection with no need to know the novel class anchor size. \ours uses foundation models to bridge the semantic gap between 3D features and texts while jointly detecting attributes, \eg, spatial relationships, motion states, \etc. 
To facilitate such research direction, we propose \data, a new dataset that supplements existing 3D object detection benchmarks with comprehensive attribute annotations. 
\ours incorporates several key innovations, including foundation model feature concatenation, prompt tuning strategies, and specialized techniques for attribute detection, including perspective-specified prompts and horizontal flip augmentation. 
Our results on both the nuScenes and Argoverse 2 datasets show that under the condition of no given anchor sizes of novel classes, \ours outperforms the state-of-the-art methods in open-vocabulary 3D object detection while successfully recognizing object attributes. Our OVAD dataset is released here\footnote{\url{https://doi.org/10.5281/zenodo.16904069}}.
\end{abstract}

\blfootnote{$^\star$ This work was done when Xinhao Xiang was an intern at MERL.}

\section{Introduction}
\label{sec:intro}

Autonomous systems require advanced 3D perception, but most rely on closed-set models limited to predefined object classes, failing with novel objects and complex scenes~\cite{PVRCNN, pointRCNN, votenet, FSD, TransFusion, DSVT, TED, OcTr, hegde2025cvprw, li2025cvprw, hegde2024eccv}. Beyond detection, understanding attributes such as motion and spatial relationships is critical. Multimodal foundation models offer promise via open-vocabulary, zero-shot learning~\cite{CLIP, CogVLM, OneLLM, GLIP}, but 3D data's sparsity, complexity, and need for class-specific anchors~\cite{Find_n_Propagate} pose challenges. Attribute detection, vital in tasks like autonomous driving, is underexplored due to limited dataset annotations~\cite{KITTI, WOD, Argoverse2} and narrow focus on classification/localization~\cite{pointRCNN, votenet, Find_n_Propagate, CoDAv2}.

\begin{figure*}[t]
\centering

\begin{minipage}[t]{0.48\textwidth}
\centering
\includegraphics[width=\linewidth]{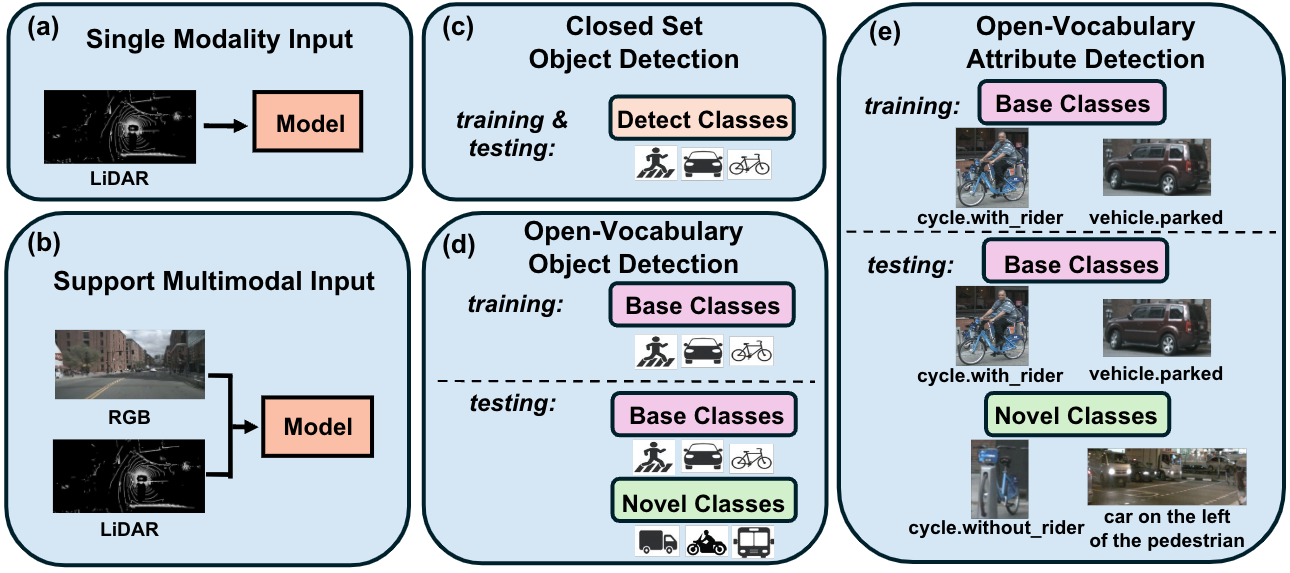}
\captionof{figure}{Relaxing the constraints of (a) single-modal (c) closed set LiDAR 3D object detection, \ours can perform (b) multimodal open-vocabulary (d) object and (e) attribute detection.}
\label{fig:intro}
\vspace{-0.25cm}
\end{minipage}
\hfill
\begin{minipage}[t]{0.5\textwidth}
\centering
\vspace*{-2.7cm}
\setlength{\tabcolsep}{4pt}
\resizebox{\linewidth}{!}{
\begin{tabular}{@{}rc@{\hspace{.3em}}c@{\hspace{.3em}}c@{\hspace{.3em}}c@{\hspace{.3em}}c@{\hspace{.3em}}c@{\hspace{.3em}}c@{\hspace{.6em}}>{\columncolor{red!10}}c}
\toprule
\multirow{2.5}{*}{\textbf{\shortstack{Object/Attribute Detection \\Conditions}}} & \multicolumn{8}{c}{\textbf{Method Categories}} \\ \cmidrule(l){2-9} 
 & $\cC_0$ & $\cC_1$ & $\cC_2$ & $\cC_3$ & $\cC_4$ & $\cC_5$ & $\cC_6$ & \ourso \\ 
\midrule
support 3D object detection & $\ccheck$ & \ccross & $\ccheck$ & $\ccheck$ & $\ccheck$  & $\ccheck$ & \ccross & $\ccheck$ \\
can detect attributes & \ccross & \ccross & \ccross & \ccross & \ccross  & $\ccheck$ & \ccross & $\ccheck$ \\
support OV object detection & \ccross & $\ccheck$ & $\ccheck$ & $\ccheck$ & $\ccheck$  & \ccross & \ccross & $\ccheck$ \\
support multi-modal input & \ccross & \ccross & $\ccheck$ & \ccross & $\ccheck$  & \ccross & \ccross & $\ccheck$ \\
need \textbf{no} novel class anchor size & \ccross & $\ccheck$ & \ccross & $\ccheck$ & $\ccheck$  & \ccheck & \ccross & $\ccheck$ \\
can detect complex events & \ccross & \ccross & \ccross & \ccross & \ccross  & \ccross & $\ccheck$ & $\ccheck$ \\
\bottomrule
\end{tabular}
} 
\captionof{table}{\ours addresses 3D object detection challenges which were formerly not fully considered in prior works (\eg, $\cC_0$: \cite{PVRCNN, pointRCNN, votenet, FSD, TransFusion, DSVT, FusionViT, 3DifFusionDet, EffiPerception}, $\cC_1$: \cite{OVR_CNN, OV_DETR, ViLD, VL_PLM, Detpro, GroundingDINO}, $\cC_2$: \cite{Find_n_Propagate}, $\cC_3$: \cite{PointCLIP_2, Object2Scene}, $\cC_4$: \cite{OpenScene, OpenShape, OV_3DETIC, OV_3DET, CoDA, FM_OV3D, OpenSight, CoDAv2, ConceptFusion}, $\cC_5$: \cite{PointNet, Frustum_PointNets, Graph_RCNN, HOV_SG, ConceptGraphs}, $\cC_6$: \cite{HMM, C3D, I3D, VideoBERT, timesformer, MMT, Action_Genome}). OV: open-vocabulary.}
\label{tab:related_works}
\end{minipage}

\vspace*{-\baselineskip}
\end{figure*}

To address these challenges, we propose Open-Vocabulary Object Detection with Attributes (\ours), a framework for open-vocabulary 3D object and attribute detection that uses foundation models (FM) for semantics while preserving 3D geometric precision. Unlike prior methods (\eg, \cite{Find_n_Propagate}), \ours detects novel classes and attributes without anchor size knowledge of the novel classes. \ours combines temporal-spatial features, complex event generation, and semantic attribute alignment, using FM features and prompt tuning to unify 3D geometry and semantics.

\ours integrates attribute detection into object detection, enabling unified recognition of novel objects and their attributes—spatial relations, motion states, and interactions. This is achieved via a complex event generation module that aligns 3D features with text in semantic space. Perspective-specific prompts and horizontal flip augmentation further improve accuracy under challenging conditions with varying viewpoints and object orientations. \ours thus delivers comprehensive scene understanding by identifying objects, their relations, and behaviors.

To boost \ours’s performance, we add two enhancements: (1) Combining FM and existing 3D detection backbone features for richer semantics understanding and precise localization; (2) Prompt tuning for task-specific FM adaptation. We also introduce two loss functions for learning novel attributes without annotations associated with them, making \ours a robust effective solution for open-vocabulary 3D object and attribute detection. To facilitate attribute detection research, we propose the Open Vocabulary Attribute Detection (\data) dataset, a benchmark built on nuScenes~\cite{nuScenes} with 84384 instances labeled across 11 attribute classes. \data is the first benchmark to include detailed annotations on spatial relations, motion states, and interactions, enabling thorough evaluation of complex scene understanding and open-vocabulary 3D attribute detection in real outdoor scenes.

Our experiments on nuScenes~\cite{nuScenes} and Argoverse 2~\cite{Argoverse2} show that \ours beats the state-of-the-art (SOTA) in open-vocabulary 3D object detection when the novel class anchor sizes are unavailable, while also detecting novel attributes. This marks a key advance in 3D scene understanding with applications like autonomous driving and robotics. Our contributions include:

\begin{enumerate}
[label=\arabic*., leftmargin=*, topsep=0pt]
\setlength\itemsep{-0.0em}
    \item We propose \ours, a novel open-vocabulary multimodal 3D object detector from multi-view input to detect complex events (including attributes) without needing novel class anchor size.
    \item We propose concatenation with foundation model features, prompt tuning strategies, two novel loss functions, and two attribute-specific techniques (perspective-specified prompt, horizontally flip augmentation), to improve the open-vocabulary object and attribute detection performance.
    \item Proposing the \data dataset for open-vocabulary attribute detection, we show that under the condition of no predefined novel class anchor sizes, \ours outperforms the SOTA methods on the open-vocabulary 3D object detection task on both the nuScenes and Argoverse 2 datasets.
\end{enumerate}

\section{Related work}
\label{sec:formatting}

Contrasting \ours with prior works in Tab.~\ref{tab:related_works}, we categorize them by properties and elaborate.

\noindent\textbf{Open-vocabulary (OV) 3D object detection.}
OV detection uses language models to classify both seen and novel classes, offering greater flexibility than traditional open-set or zero-shot learning. While most 2D OV methods (\eg, $\cC_1$) use pretrained vision-language (VL) models like CLIP~\cite{CLIP} and GLIP~\cite{GLIP}, conventional 3D detectors ($\cC_0$) rely on closed-set supervision. Recent works ($\cC_3$, $\cC_4$) adapt VL features to 3D perception via multi-modal embeddings~\cite{ConceptFusion}, but struggle to encode spatial cues from sparse point clouds and focus mainly on indoor scenes. \ours overcomes these limits in diverse and dynamic outdoor settings with complex spatial relationships.

\noindent\textbf{Complex event extraction} captures object relationships, temporal dynamics, and context, unlike basic object detection.  Events like “person behind car” or “person sitting” involve multiple objects or attributes. Prior methods ($\cC_5$) use handcrafted features or models like HMMs~\cite{HMM}, CNN-~\cite{C3D, I3D}, RNN-~\cite{LSTM}, and Transformer-based~\cite{VideoBERT, timesformer} models, but struggle with generalization and efficiency. Multimodal~\cite{MMT} and graph-based~\cite{Action_Genome} methods add richer context but face alignment or manual setup issues. \ours is the first to detect complex events in OV 3D settings by jointly predicting objects and attributes.

\noindent\textbf{Attribute detection in 3D outdoor scenes.}
Detecting attributes like motion and spatial attributes in 3D outdoor scenes is vital but hard due to sparse sensor data and complex scenes. Prior methods ($\cC_4$) extend detectors to predict attributes~\cite{PointNet} or fuse modalities~\cite{Frustum_PointNets}, but need fine-grained cues or precise calibration. Graph-based approaches~\cite{Graph_RCNN, ConceptGraphs} need manual graph design and struggle with 3D sparsity. \ours uses foundation models~\cite{OneLLM, CogVLM} to incorporate semantics, enabling open-vocabulary attribute detection with better generalization.

\section{Our proposed method --- \ours}
\label{sec:method}

\subsection{Framework overview}

\begin{figure*}[t]
    \centering
    \includegraphics[width=.92\linewidth]{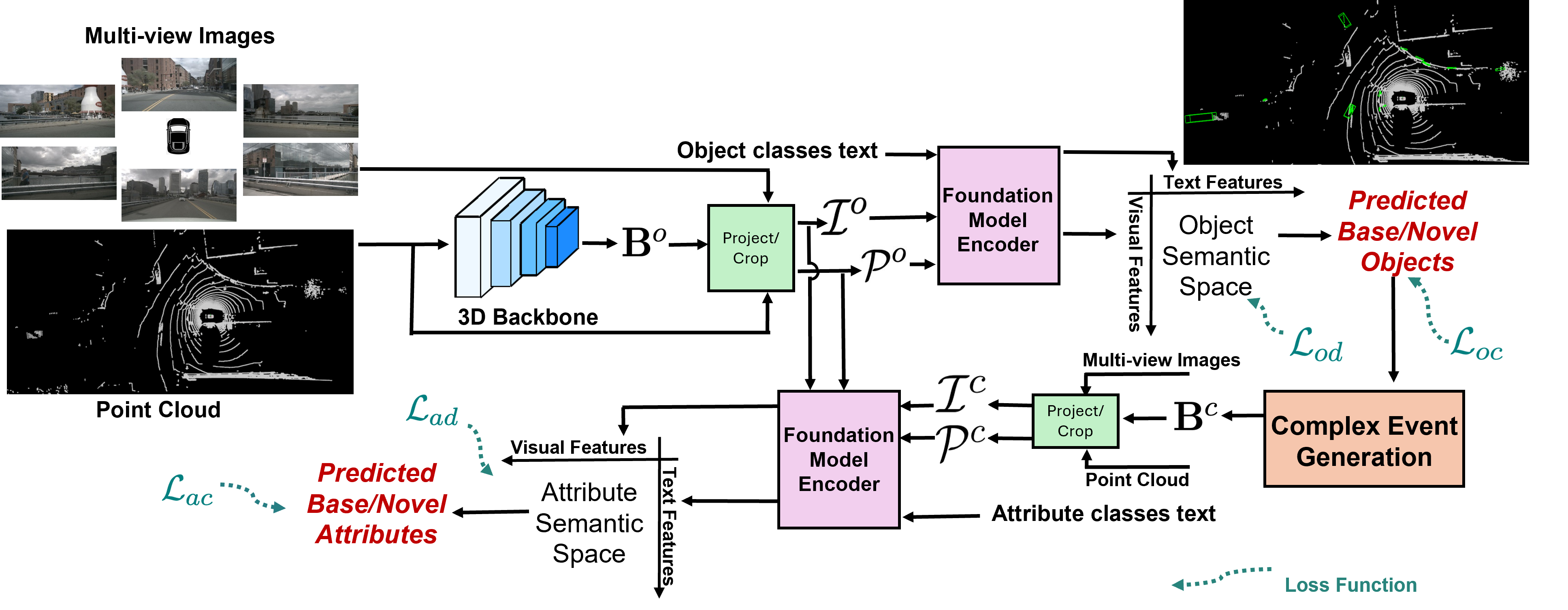}
    \vspace{-1mm}
    \caption{Our proposed \ours framework combines information from multi-view images and point clouds, align object and attribute text with vision features using a common foundation model encoder in order to discover and localize complex open-vocabulary events that include multiple objects and attributes. We show the figure using information from a single time instant for simplicity. In our method, we aggregate information over multiple temporal instances enabling better complex event discovery involving motion attributes.
    }
    \label{fig:framework1}
    \vspace{-.5em}
\end{figure*}

We define the OV detection setup with object and attribute labels available for base classes $\CC^b$ and $\CC^{ba}$ during training. At inference, the goal is to detect both base and novel object classes $\CC^n$ and attributes $\CC^{na}$. Complex events refer to (a) an object with an attribute or (b) two objects with a spatial relation.
As shown in Fig.~\ref{fig:framework1}, \ours is the first OV 3D detector to jointly recognize objects and complex events. It processes 3D point clouds and multi-view images via a spatiotemporal extractor (STE) to produce single-object proposals (Sec.~\ref{sec:single}), and builds on 3DETR~\cite{3detr} for transformer-based localization and classification.
We enable OV detection by aligning vision, point cloud, and text features using a frozen OneLLM~\cite{OneLLM} model, allowing recognition of novel classes in the FM’s semantic space. The \complex module (Sec.~\ref{sec:complex}) proposes spatially related object pairs for complex event detection, which are matched in attribute space for novel attribute recognition. \ours is trained end-to-end with joint losses for both tasks (Sec.~\ref{sec:optimization}).

\subsection{Single-object proposal}
\label{sec:single}

\noindent\textbf{Single-object proposal extraction.} We begin with a set of base-class objects in the training dataset with ground truth 3D object bounding box: $\Db^b=\left\{o_j=\left(c_j, B_j\right) \mid c_j \in \CC^b\right\}$,
where $c_j$ is a class in the set of the base object classes ($\CC^b$) and $B_j$ is the corresponding ground truth bounding box. 
We train an initial class-agnostic 3D object proposer $f_{\text{det}}$ using $\Db^b$ by minimizing an object box regression loss inspired by 3DETR~\cite{3detr}.
By focusing exclusively on objectness score prediction and box regression, we avoid the limitations of class-specific training that have been shown to hinder novel object detection in OV 2D detection~\cite{survey}.
The trained $f_{\text{det}}$ generates hidden features $\Fb_{det}$ and object proposals $\Bb^o$ which is characterized by objectness scores and precise 3D localization parameters.

\noindent\textbf{Single-object proposal generation.} $\Bb^o$ guides the generation of single-object instances from multimodal visual inputs (point cloud $\cP$ and multi-view images $\cI$) for subsequent text-visual alignment and class prediction.
We crop the object instances from the 3D boxes $\cP^o$ in $\cP$ via $\cP^o=\operatorname{Crop}\left(\Bb^o, \ \cP\right)$.
For $\cI$, we first project $\Bb^o$ onto the 2D image plane using the camera matrices $M$:
$
\Bb_{2D}^{o}=\operatorname{Proj}\left(\Bb^{o}, \ M\right),
$
and then we crop the corresponding objects in the image using $\cI^{o}=\operatorname{Crop}\left(\Bb_{2D}^{o}, \ \cI\right).$
All cropped image features are concatenated together with an additional encoding representing view direction.

\noindent\textbf{Novel object class discovery.} 
To recognize novel object classes, we use the frozen OneLLM~\cite{OneLLM} to align image and point cloud features of each object proposal $B_j^o$—denoted by $V_j^O = [\cI_j^o; \cP_j^o]$—with a super-class vocabulary $T^O$ containing both base ($\CC^b$) and novel ($\CC^n$) classes. The prediction distribution over $C$ classes is:
$\Pb_j^{O} =\left\{p_{j, 1}^O, \ p_{j, 2}^O, \ \ldots \ , \ p_{j, C}^O\right\} =\operatorname{Softmax}\left(V_j^{O} \cdot \Fb_{T^O}\right)$,
where $\Fb_{T^O}$ is the textual embedding of $T^O$, $\cdot$ denotes the dot product, and the distribution $\Pb^O_j$ serves as the OV semantic priors. The final predicted class is $c_j^* = \arg\max_c \Pb_j^O$, and the resulting single-object detections are:
$
\Db^s = \left\{\left(c_j^*, \ B_j^o\right)\mid c_j^* \in \CC^b\cup\CC^n\right\}
$.
We classify $B_j^o$ as a novel object if it meets:
\fontsize{9.9pt}{9.9pt}
\begin{equation}
\label{eq:8}
\begin{gathered}\Ob^{disc}=\left\{B_j^o \mid \forall B_i^b \in \Bb^b, \  \operatorname{IoU}_{3 D}\left(B_j^o, \ B_i^b\right)<\theta^b, \ Q_j^o>\theta^o,\right. \\ \left.p_{j, c^*_j}^O>\theta^s, \ B_j^o \in \Bb^o, \ c_j^* \notin \CC^b\right\},\end{gathered}
\end{equation}
where $\Bb^b$ is the set of proposals classified as base classes, $Q_j^o$ is the 3D objectness score, and $\theta^b=0.2$, $\theta^o=0.8$, $\theta^s=0.5$ are thresholds for IoU, objectness, and semantic confidence.

\noindent\textbf{Detector improvement strategy.} We propose two key enhancements for $f_{\text{det}}$ to improve OV detection performance: (1) Feature Fusion with FM: We augment $f_{\text{det}}$'s encoded features with visual features extracted from OneLLM~\cite{OneLLM}. This multi-dimensional feature fusion strategy enriches the detector's input representation, enabling more comprehensive learning from limited data. (2) Prompt Tuning Integration: We incorporate learnable visual prompts at the FM's input layer, facilitating task-specific adaptation of the pretrained FM.

\subsection{Complex event generation (\complex)}
\label{sec:complex}

\ours extends beyond traditional OV 3D object detection to be able to detect complex events via jointly predicting objects and their attributes.
We consider some of the most critical outdoor scene attributes, including spatial relationships between objects, motion states, and human-traffic participant interactions. 
To this end, different from conventional methods focusing only on single-object detection, \ours includes a novel complex event proposal generation method to extract complex inter-object contextual knowledge which is crucial to support downstream complex event detection.

\noindent\textbf{Complex event visual proposal generation.}
\ours addresses three critical outdoor scene attributes: spatial relationships, motion states, and human-traffic participant interactions. 
Effective attribute detection needs both spatial and temporal context beyond single objects.
The spatial attribute detection relies on relative positional and orientational relationships between objects, while temporal attribute (\eg, motion state) detection relies on the temporal information across multiple timestamps.
To meet these needs, we construct complex event proposals by concatenating: (1) Non-spatial attribute features: Temporal sequence of single-object proposals. (2) Spatial attribute features: Proposals generated from two nearby single-object proposals.
We define $\Db^o$ as the set of detected single-object proposals: $\Db^o=\left\{\left(c^o_j, \: B^o_j\right) \mid c^o_j \in \CC^b\cup\CC^n\right\}$. 
To generate non-spatial attribute proposals $\Bb^n$, we concatenate current single-object proposals with those generated in the past $T$ timestamps in total, to incorporate temporal sequence information into the proposals (\ie $\Bb^n=\left(\Bb^o, \ \Bb^{o-1}, \ \dots \ , \ \Bb^{o-T}\right)$).
The spatial attribute proposals $\Bb^s$ are generated by combining nearby single-object proposals:
\begin{equation}
\Bb^s=\left\{B_{ij}^s=\left(\operatorname{Comb}\left(B_i^o, \ B_j^o\right) \mid \operatorname{Dist}\left(B_i^o, \ B_j^o\right)\leq\theta^d\right)\right\},
\end{equation}
where $\theta^d$ is the distance threshold of 15 meters. The $\operatorname{Comb}$ operation creates a larger two-object proposal $B_{ij}^s$ by merging selected single-object proposals $B_i^o$ and $B_j^o$. $B_{ij}^s$'s spatial extent is defined by:
\fontsize{9.6pt}{9.6pt}
\begin{equation}
\left[
\begin{array}{c}
x_{min} = \operatorname{Min}\left(B_{i x}^o, \  B_{j x}^o \right),\quad x_{max}=\operatorname{Max}\left( B_{i x}^o, \ B_{j x}^o\right), \\[10pt]
y_{min} = \operatorname{Min}\left(B_{i y}^o, \  B_{j y}^o \right),\quad y_{max} = \operatorname{Max}\left( B_{i y}^o, \ B_{j y}^o\right), \\[10pt]
z_{min} = \operatorname{Min}\left(B_{i z}^o, \ B_{j z}^o \right),\quad z_{max} = \operatorname{Max}\left( B_{i z}^o, \ B_{j z}^o\right)
\end{array}
\right]
\label{eq:3}
\end{equation}
\normalsize
   
The final set of generated complex event visual proposals $\Bb^c$ is the concatenation of the two feature groups (\ie $\Bb^c =\Bb^n \cup \Bb^s$.)
Following the same $\operatorname{Crop}$ and $\operatorname{Proj}$ operations in Sec.~\ref{sec:single},
we extract corresponding visual proposals in image ($\cI^c$) and point cloud ($\cP^c$) modalities. We perform data augmentation by horizontal flipping (\ie, $\cI_{flip}^c = \operatorname{Flip}(\cI^c)$) for more robust detection. 
For each proposal, we perform inference by using $\cI^c$ and $\cI_{flip}^c$ in turn as input.
We average the prediction scores from these two to get the final prediction result.

\noindent\textbf{Complex event text proposal generation.}
For each complex visual proposal $B^c$, we generate the corresponding text which can be fed into OneLLM's text encoder.
For non-spatial attribute text proposals $c^n$, the text is directly from the predicted object class (\ie, $c_j^n=\TT(c^*_j)\TT(\operatorname{NSA})$, where $\TT(.)$ is the function translating the input class label or spatial attribute to the corresponding text), $\operatorname{NSA}$ is one of the non-spatial attributes. 
For spatial attribute text proposals $c^s$, their text are generated based on the constituent single proposals and their relative spatial configuration.
Specifically, if $B_i^o$ is combined with $B_j^o$, its corresponding text is: 
\fontsize{9.6pt}{9.6pt}
\begin{equation}
\label{eq:11}
\begin{aligned}
c_{ij}^s = ``\text{From the perspective of} \
\TT(c_j^o), \ \TT(c_i^o) \ 
 \TT(\operatorname{SA}) \ \TT(c_j^o)."
\end{aligned}
\end{equation}
\normalsize
where $\operatorname{SA}$ is one of the four spatial attributes: \textit{in front of}, \textit{behind}, \textit{on the left of}, \textit{on the right of}. The classes are derived from relative coordinates of the two single proposals by mathematical definitions, making them inherently objective.
The perspective-based prefix ensures unique and distinguishable text features while establishing clear spatial relationships.
The final set of generated complex event text proposals $c^c$ is defined as: $c^c =c^n \cup c^s$. Overall, the set of generated complex events are denoted as: $\Db^c = \left\{\left(c_j^c, \ B_j^c\right)\mid c_j^c \in \CC^{ba}\cup\CC^{na}\right\}$.

\noindent\textbf{Novel attribute class discovery.} 
We use OneLLM~\cite{OneLLM} to align these proposals from image, point and text modalities.
Firstly, we generate visual feature $V^{A}$ by concatenating from image encoder feature of $\mathcal{I}^c$ or $\mathcal{I}_{flip}^c$, and point cloud encoder feature of $\mathcal{P}^c$ from OneLLM.
Then, we use OneLLM's text encoder to extract the text proposals $c^c$ to get the text features $\Fb_{T^A}$.
$V^{A}$ and $\Fb_{T^A}$ are then be aligned in the attribute semantic space where we compute the distance between them: $\Pb_j^{A} =\left\{p_{j, 1}^A, \ p_{j, 2}^A, \ \ldots \ , \ p_{j, E}^A\right\} =\operatorname{Softmax}\left(V_j^{A} \cdot \Fb_{T^A}\right)$.
Where $E$ is the total number of attribute classes. The distribution $\Pb^A_j$ serves as the OV semantic priors. 
The final predicted attribute class $e_j^*$ for each $B_j^c \in \Bb^c$ is decided by the maximum probability among $\Pb_j^{A}$, \ie, $e_j^* = \text{argmax}_e\Pb_j^A$. 
The set of detected complex event proposals is denoted as: $\Db^c = \left\{\left(e_j^*, \ B_j^c\right)\mid e_j^* \in \CC^{ba}\cup\CC^{na}\right\}$.
We determine whether a complex event proposal involves a novel attribute or not by the following criteria:
\begin{equation}
\begin{gathered}\Ab^{disc}=\left\{B_j^c \mid \forall B_i^{ba} \in \Bb^{ba}, \  \operatorname{IoU}_{3 D}\left(B_j^c, \ B_i^{ba}\right)<\theta^b\right. ,\\ \left.p_{j, e^*}^A>\theta^a, \ B_j^c\in\Bb^{c}, \ e^* \notin \CC^{ba}\right\},\end{gathered}
\end{equation}
where $\Bb^{ba} \in \Bb^c$ denotes attribute proposals in $\Bb^c$ predicted as one of the base attribute classes.
$\theta^a$ is the threshold for attribute semantic scores set to 0.5.
$\CC^{ba}$ is the set of base attribute classes.

\subsection{Overall optimization}
\label{sec:optimization}

\noindent\textbf{Open-vocabulary object losses.} To transfer knowledge from LiDAR to images, following~\cite{CoDAv2}, we enforce $V_j^O$ and $\Fb_{det}$ to be the same by using a class-agnostic L1 loss to minimize their feature distance: $\cL_{od} = \sum_{j=1}^N || V_j^O - \Fb_{det} ||_1$,
where $N$ is the number of object proposals.
$\cL_{od}$ effectively reduces cross-modal gaps, enhancing feature alignment across diverse scenes, including background regions. $\mathcal{L}_{od}$ is independent of class annotations, as it needs no ground-truth box class labels.

Like~\cite{CoDAv2}, we also use a loss to promote discriminative classification by maximizing the matching score of the ground-truth class while minimizing scores for other classes: $\cL_{oc} = \sum_{j=1}^N f\left(B_j^{disc}, \: \mathbf{B}^{b}\right) \cdot C E\left(\Pb_j^{disc}, \ h_{j}^O\right)$,
where $B_j^{disc}$ is the $j$-th object proposal of $\Ob^{disc}$. $\Pb_j^{disc}=\operatorname{Softmax}\left(V_j^{disc} \cdot \Fb_{T^O}\right)$, where $V_j^{disc}$ is the visual features of $B_j^{disc}$.
$CE(.)$ denotes the cross-entropy loss. 
The function $f(x)$ is to check if $B_j^{disc}$ is within $\mathbf{B}^{b}$, returning 1/0 when the answer is yes/no.
$\Pb_j^{disc}$ is the probability of $\Ob^{disc}$.
$h^O_j$ is the ground truth one-hot vector for $B^{disc}_j$.

\noindent\textbf{Open-vocabulary attribute losses.}
Similar to object losses, we align $V_j^A$ with $\Fb_c$ to transfer knowledge from LiDAR to images, where $\Fb_{c}$ is the 3D backbone features of $\Bb^c$.
We enforce $V_j^A$ and $\Fb_c$ to be the same via a class-agnostic L1 loss to minimize their feature distance: $\cL_{ad} = \sum_{j=1}^N ||V_j^A - \Fb_c||_1$.

In addition, we propose to use a contrastive loss to ensure correct base attribute classification by maximizing the matching score of the ground-truth attribute class while minimizing the scores for other attribute classes: $\cL_{ac} = \sum_{j=1}^N g\left(A_j^{disc}, \: \Bb^{ba}\right) \cdot C E\left(\mathbf{P}_{j}^{disc, a}, \ h_{j}^A\right)$,
where $A_j^{disc}$ is the $j$-th complex event proposal of $\Ab^{disc}$.
$\Pb_j^{disc, a}=\operatorname{Softmax}\left(V_j^{disc, a} \cdot \Fb_{T^A}\right)$, where $V_j^{disc, a}$ is the visual features of $A_j^{disc}$.
The function $g(x)$ is to check if $A_j^{disc}$ is within $\mathbf{B}^{ba}$, returning 1/0 when the answer is yes/no. 
$h_j^A$ is the one-hot ground truth attribute vector for $A_j^{disc}$.

These aforesaid four losses jointly improve feature alignment, making the 3D features of novel objects/attributes more discriminative, thus enhancing the model's ability to detect novel objects/attributes.
The final loss function $\cL$ is defined as: $\cL = w_{od}\cL_{od} + w_{oc}\cL_{oc} + w_{ad}\cL_{ad} + w_{ac}\cL_{ac}$,
where $w$'s represent the weights balancing each loss to ensure comparable ranges.

\section{The \data dataset}
\label{sec:dataset}

\ours is the first method to perform complex event detection in OV 3D obejct detection by jointly detecting objects and key outdoor attributes: spatial relations, motion state, and presence of people among traffic participants. Existing datasets lack full annotations—\eg, nuScenes~\cite{nuScenes} includes motion and presence of people but not spatial attributes. To fill this gap, we propose a novel attribute dataset, \data, for comprehensive attribute training and detection evaluation.

Built on nuScenes, \data retains its attribute annotations. From 28,130 nuScenes time instances, 5,000 were sampled, yielding 170,149 object annotations, filtered to 84,384 annotations across 10 target classes. To create spatial attribute annotations, we selected pairs of object annotations with a distance within $[0\text{m}, 15\text{m}]$: $\Bb^{\text{\data}}=\left\{ B_{ij}^{\text{\data}}=\left(\operatorname{Comb}\left(B_i, \ B_j\right) \mid \operatorname{Dist}\left(B_i, \ B_j\right)\leq15\text{m}\right)\right\}$.
The $\operatorname{Comb}$ operation creates a larger two-object proposal $B_{ij}^{\text{\data}}$ by merging the selected two nearby ground truth single-object proposals $B_i$ and $B_j$. The spatial extent of $B_{ij}^{\text{\data}}$ is defined in a similar way as Eq.~\ref{eq:3}. The 15-meter spatial threshold follows real-world traffic constraints and practical sensor limitations to ensure reliable sampling.
For the ground-truth label of $B^{\text{\data}}$ (\ie, $c_{ij}^{\text{\data}}$), its text is generated based on the constituent single proposals and their relative spatial configuration. Specifically, if $B_i$ is combined with $B_j$, its corresponding text is: $c_{ij}^{\text{\data}} = \TT(B_i) \ \TT(\operatorname{SA}) \ \TT(B_j)$.

\section{Experiments}
\label{sec:exp}

\noindent\textbf{Dataset \& metrics.}
We evaluate on nuScenes~\cite{nuScenes} and Argoverse 2~\cite{Argoverse2} using mean Average Precision (mAP), nuScenes Detection Score (NDS) for object detection, success rate (SR) for attribute detection, and $\text{AP}_\text{N}$ (mean AP computed only over novel classes). In nuScenes settings, we use 10 object classes, 7 non-spatial, and 4 spatial attribute classes from \data; in Argoverse 2 settings, we use 8 object classes for consistency with nuScenes. Our OV settings for object and attribute detection are detailed in Tab.~\ref{tab:ov_object} and~\ref{tab:ov_a}, respectively. Additional details about our dataset settings, metrics settings, vocabulary settings, and implementation are in the supplement.

\begin{figure*}[t]
\centering

\begin{minipage}[t]{0.56\textwidth}
\centering
\resizebox{.9\linewidth}{!}{
\begin{tabular}{@{}c@{}c@{}c@{}}
\toprule
dataset setting & base object class & novel object class \\
\midrule
$N_{b6n4}$  & \makecell{Car, Construction vehicles, Trailer, \\ Barrier, Bicycle, Pedestrian} & \makecell{Truck, Bus, \\ Motorcycle, Traffic cone} \\
\midrule
$N_{b3n7}$  & \makecell{Car, Bicycle, Pedestrian} & \makecell{Construction vehicles, Trailer, Barrier, \\ Truck, Bus, Motorcycle, Traffic cone} \\
\midrule
$N_{b0n10}$  & \makecell{$\varnothing$} & \makecell{Car, Construction vehicles, Trailer, \\ Barrier, Bicycle, Pedestrian Truck, \\ Bus, Motorcycle, Traffic cone} \\
\midrule
$A_{b4n4}$ & \makecell{Regular Vehicle, Trailer, \\  Bicycle, Pedestrian} & \makecell{Truck, Bus, \\Motorcycle, Construction cone} \\
\bottomrule
\end{tabular}
}
\captionof{table}{
For fair comparison, we follow~\cite{Find_n_Propagate} and evaluate on nuScenes using 3 OV settings ($N_{b6n4}$, $N_{b3n7}$, $N_{b0n10}$). We select the same object classes in Argoverse 2 and split them into base and novel classes ($A_{b4n4}$) for consistency with nuScenes.
}
\label{tab:ov_object}
\vspace{-.3cm}
\end{minipage}
\hfill
\begin{minipage}[t]{0.43\textwidth}
\centering
\vspace{-1.0cm}
\resizebox{\linewidth}{!}{
\begin{tabular}{@{}c@{}cc@{}}
\toprule
dataset setting & base attribute class & novel attribute class \\
\midrule
\textit{\data} & \makecell{\textcolor{teal}{with rider}, \textcolor{violet}{sitting lying down}, \\ \textcolor{brown}{parked}, in front of, behind} & \makecell{\textcolor{teal}{without rider}, \textcolor{violet}{standing}, \\ \textcolor{brown}{moving}, on the left of} \\
\bottomrule
\end{tabular}
}
\captionof{table}{The OV settings on attribute classes in nuScenes. Our \data dataset includes all the attributes in nuScenes and 4 spatial classes. The attributes colored with \textcolor{teal}{teal}/\textcolor{violet}{violet}/\textcolor{brown}{brown} are the attributes exclusively associated with \textcolor{teal}{cycle}/\textcolor{violet}{pedestrian}/\textcolor{brown}{vehicle} classes. We included all providing attribute annotations in nuScenes.}
\label{tab:ov_a}
\vspace{-.3cm}
\end{minipage}
\end{figure*}

\subsection{Experimental results}

\begin{table*}[t]
\centering
\resizebox{\linewidth}{!}{
\begin{tabular}{@{}ccccccccccccc@{}}
\toprule[1pt]
\multirow{2.5}{*}{method} &\multirow{2.5}{*}{CFM} &\multirow{2.5}{*}{\shortstack{prompt\\tuning}} & \multirow{2.5}{*}{\shortstack{need \textbf{no} predefined anchor\\size for each novel class?}} & \multicolumn{3}{c}{$N_{b6n4}$}  & \multicolumn{3}{c}{$N_{b3n7}$} & \multicolumn{3}{c}{$N_{b0n10}$}  \\
\cmidrule(l){5-7} \cmidrule(l){8-10} \cmidrule(l){11-13}
& & & & mAP & NDS & $\text{AP}_{\text{N20}}$ & mAP & NDS & $\text{AP}_{\text{N20}}$ & mAP & NDS & $\text{AP}_{\text{N20}}$\\
\midrule[.75pt]
Find $\mathrm{n}^{\prime}$ Propagate~\cite{Find_n_Propagate} &\ccross &\ccross &\ccross & 44.95 & 47.87 & 33.65 & 37.38 & 40.28 & 18.46 & N/A & N/A & 16.72 \\
\midrule
CoDAv2~\cite{CoDAv2} &\ccross &\ccross &$\ccheck$ & 27.35 & 29.48 & 12.63 & 18.73 & 20.14 & 8.74 & 4.32 & 5.82 & 1.37  \\
\ours &\ccross &\ccross &$\ccheck$ & 30.24 & 31.54& 14.24 & 20.46 & 21.83 & 9.31  & 4.57 & 6.96 & 2.16 \\
\ours &$\ccheck$ &$\ccheck$ &$\ccheck$ & \textbf{32.25} & \textbf{31.85} & \textbf{14.72} & \textbf{21.03} & \textbf{22.14} & \textbf{10.23} & \textbf{4.70} & \textbf{7.05} & \textbf{2.39} \\
\bottomrule[1pt]
\end{tabular}
}
\caption{OV object detection results on nuScenes. Acronym: CFM: concatenating foundation model features.
}
\label{tab:nuscenes_result}
\vspace{-.5em}
\end{table*}

\begin{figure*}[t]
\centering

\begin{minipage}[t]{0.54\textwidth}
\centering
\vspace{-1.6cm}
\resizebox{.8\linewidth}{!}{
\begin{tabular}{ccccc}
\toprule
method & CFM & prompt tuning & mAP & $\text{AP}_{\text{N20}}$ \\
\midrule
\ours & \ccross & \ccross & 17.24 & 8.23 \\
\ours & $\ccheck$ & $\ccheck$ & \textbf{27.43} & \textbf{12.34} \\
\bottomrule
\end{tabular}
}
\captionof{table}{Performance comparison of different methods on Argoverse 2 under the $A_{b4n4}$ setting. Acronyms: CFM: concatenating foundation model features.}
\label{tab:argo_result}
\resizebox{.93\linewidth}{!}{
\begin{tabular}{@{}c@{\hspace{0.3em}}c@{\hspace{0.5em}}c@{\hspace{0.5em}}c@{\hspace{0.5em}}c@{\hspace{0.5em}}c@{\hspace{0.3em}}c@{\hspace{0.3em}}c@{}}
\toprule
\multirow{2}{*}{method} & \multirow{2}{*}{CFM} & \multirow{2}{*}{PT} & \multirow{2}{*}{mAP} & \multirow{2}{*}{NDS} & \multirow{2}{*}{$\text{AP}_{\text{N20}}$} & \multirow{2}{*}{\shortstack{SR (\%)\\(AD only)}} & \multirow{2}{*}{\shortstack{SR (\%)\\(AD \& OD)}} \\
& & & & & & & \\
\midrule
\multirow{3}{*}{\ours} & \ccross & \ccross & 30.24 & 31.54 & 14.24 & 16.35 & 4.23 \\
& $\ccheck$ & \ccross & 31.34 & 31.63 & 14.42 & 22.74 & 5.56 \\
& $\ccheck$ & $\ccheck$ & \textbf{32.25} & \textbf{31.85} & \textbf{14.72} & \textbf{25.90} & \textbf{6.77} \\
\bottomrule
\end{tabular}
}
\captionof{table}{Both the object and attribute detection results as well as the ablation study of our framework on the nuScenes dataset, under $N_{b6n4}$ object setting and \textit{\data} attribute setting. Acronyms: SR: success rate; AD: attribute detection; OD: object detection; CFM: concatenating foundation model features; PT: prompt tuning.}
\label{tab:promt_fm}
\end{minipage}
\hfill
\begin{minipage}[t]{0.45\textwidth}
\centering
\includegraphics[width=\linewidth]{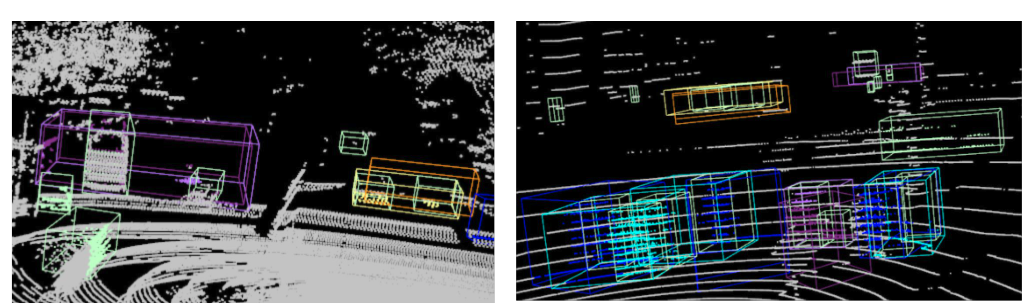}
\caption{Two qualitative results of \ours for 3D complex event detection in nuScenes dataset. All ground truth annotations of single object are rendered in \textcolor{green}{light green}. The ground truth annotations are rendered in \textcolor{pink}{light purple}/\textcolor{yellow}{yellow}/\textcolor{cyan}{light blue} for the 
\textcolor{violet}{car-car}/\textcolor{orange}{pedestrain-pedestrain}/\textcolor{blue}{others} complex events, the predicted bounding boxes are rendered in \textcolor{violet}{purple}/\textcolor{orange}{oragne}/\textcolor{blue}{blue} for the \textcolor{violet}{car-car}/\textcolor{orange}{pedestrain-pedestrain}/\textcolor{blue}{others} complex events. Examples of \textcolor{violet}{car-car}/\textcolor{orange}{pedestrain-pedestrain}/\textcolor{blue}{others} complex events can be: \textcolor{violet}{a car in front of the car}/\textcolor{orange}{a pedestrain on the left of the pedestrain}/\textcolor{blue}{a cyclist behind a car}.
}
\label{fig:qualitative2}
\end{minipage}
\vspace{-.5em}
\end{figure*}

\noindent\textbf{3D open-vocabulary object detection.}
Tab.~\ref{tab:nuscenes_result} shows the results of OV 3D object detection on nuScenes across three experimental settings. 
Prior works~\cite{OV_3DETIC, OV_3DET, OpenShape, Object2Scene} focus on indoor datasets, leaving Find n' Propagate~\cite{Find_n_Propagate} as the only outdoor 3D OV baseline. To expand comparisons, we adapt CoDAv2~\cite{CoDAv2} for nuscenes datasets by modifying only the dataloader.
The low baseline score reflects the significant difficulty of the task. The result shows that \ours outperforms CoDAv2 when novel class anchor size is unavailable, and that by concatenating FM features (CFM) and prompt tuning, \ours outperforms CoDAv2 in mAP for the $N_{b6n4}$, $N_{b3n7}$, and $N_{b0n10}$ settings, respectively. In the $N_{b0n10}$ setting (detecting completely unseen classes), our OVODA achieves an \textbf{11.4\%} performance improvement.
Although Find n' Propagate~\cite{Find_n_Propagate} performs well in its original setting, it needs predefined per-class anchor sizes to decode box geometry for novel classes ($\cC_2$ in Tab.~\ref{tab:related_works}), which we argue is an impractical and unfair advantage in the OV 3D object detection task. 
In contrast, \ours reaches competitive performance with no such constraints, showing greater flexibility and applicability.
Tab. \ref{tab:argo_result} shows \ours's result on Argoverse 2. We only use the degraded version of \ours (without CFM and prompt tuning) as the baseline because the degraded \ours already outperforms CoDAv2 on nuScenes. Tab.~\ref{tab:nuscenes_result} and~\ref{tab:argo_result} show \ours's generalizability and adaptability across different datasets. We show qualitative comparison in the supplement.

\noindent\textbf{Complex event detection.}
\ours outperforms other OV 3D detectors by jointly predicting objects and attributes for complex event detection. Tab.~\ref{tab:promt_fm} shows the results under the $N_{b6n4}$ object and \textit{\data} attribute settings. We evaluate in two modes: using the predicted objects (the last column) and ground-truth objects (the second to the last column) for isolated attribute evaluation. \ours shows strong performance in mAP, NDS, AP$_\text{N20}$, and success rates (SR), with the concatenation of FM features (CFM) and prompt tuning (PT) yielding consistent gains. Since foundation models are frozen, \ours could reach 6.77\% SR for full pipeline and runs at \textbf{27} FPS on an NVIDIA RTX A6000, confirming real-time capability. Qualitative results are in Fig.~\ref{fig:qualitative2} and the supplement.

\subsection{Ablation study}

\noindent\textbf{Foundation model augmentation \& prompt tuning.}
\begin{wraptable}{R}{0.5\textwidth}
\centering
\resizebox{0.5\textwidth}{!}{
\begin{tabular}{@{}c@{\hspace{0.3em}}c@{\hspace{0.5em}}c@{\hspace{0.5em}}c@{\hspace{0.5em}}c@{\hspace{0.3em}}c@{\hspace{0.3em}}c@{}}
\toprule
\multirow{2}{*}{method} &\multirow{2}{*}{\shortstack{foundation\\model}} & \multirow{2}{*}{mAP} &\multirow{2}{*}{NDS} &\multirow{2}{*}{$\text{AP}_{\text{N20}}$} & \multirow{2}{*}{\shortstack{SR (\%)\\(AD only)}} & \multirow{2}{*}{\shortstack{SR (\%) (for both\\AD \& OD)}} \\
& & & & & &\\
\midrule
\multirow{3}{*}{\ours} & CLIP~\cite{CLIP} & 21.93 & 21.54 & 11.45 & 16.20 & 3.22 \\
 & CogVLM~\cite{CogVLM} & 25.34 & 26.81 & 12.84 & 19.84 & 5.39 \\
 & OneLLM~\cite{OneLLM} & \textbf{32.25} & \textbf{31.85} & \textbf{14.72} & \textbf{25.90} & \textbf{6.77} \\
\bottomrule
\end{tabular}
}
\caption{Ablation study using different foundation models on nuScenes ($N_{b6n4}$/\textit{\data} for object/attribute detection settings). Acronyms: SR: success rate; AD: attribute detection; OD: object detection.}
\label{tab:ablation_attribute_model}
\end{wraptable}
We evaluate the impact of FM feature concatenation (CFM) and prompt tuning (PT) through the ablation study in Tab.~\ref{tab:promt_fm}, where both CFM and PT contribute to \ours's final performance. 
CFM enhances feature representation by using rich semantic embeddings, while PT enables better task adaptation. This synergistic effect is particularly evident in the second to the last column (attribute detection only), where \ours with CFM and PT yields a 9.55\% absolute gain in SR over the degraded \ours without CFM and PT.

\noindent\textbf{Adapting attribute detection model.}
We evaluate the impact of different FMs (CLIP~\cite{CLIP}, CogVLM~\cite{CogVLM}, and OneLLM~\cite{OneLLM}) on \ours's performance. OneLLM provides FM encoders in text, image, and point cloud modalities for \ours to use, while CLIP and CogVLM only provide FM encoders in text and image modalities.
\begin{wraptable}{R}{0.5\textwidth}
\centering
\resizebox{0.5\textwidth}{!}{
\begin{tabular}{@{}c@{\hspace{0.3em}}c@{\hspace{0.5em}}c@{}c@{\hspace{0.3em}}c@{\hspace{0.5em}}c@{\hspace{0.5em}}c@{\hspace{0.3em}}c@{\hspace{0.3em}}c@{}}
\toprule
\multirow{2}{*}{method} & \multirow{2}{*}{PSP} & \multirow{2}{*}{HFA} &\multirow{2}{*}{FM} & \multirow{2}{*}{mAP} & \multirow{2}{*}{NDS} & \multirow{2}{*}{$\text{AP}_{\text{N20}}$} & \multirow{2}{*}{\shortstack{SR (\%)\\(AD only)}} & \multirow{2}{*}{\shortstack{SR (\%) (for both\\AD \& OD)}} \\
& & & & & & &\\
\midrule
\multirow{5}{*}{\ours} &\ccross &\ccross &CLIP & 21.93 & 21.54 & 11.45 & 16.20 & 3.22\\
 &$\ccheck$ &\ccross &CLIP & 22.34 & 24.35 & 11.94 & 16.83 & 3.49\\
 &\ccross &$\ccheck$ &CLIP & 22.46 & 23.43& 12.32& 17.43& 4.03\\
 &$\ccheck$ &$\ccheck$ &CLIP & 24.37& 25.83& 13.02& 19.42 &4.92\\
 &$\ccheck$ &$\ccheck$ &OneLLM & \textbf{32.25} & \textbf{31.85} & \textbf{14.72} & \textbf{25.90} & \textbf{6.77} \\
\bottomrule
\end{tabular}
}
\caption{Ablation study of using foundation models (FM) (CLIP~\cite{CLIP} \& OneLLM~\cite{OneLLM}) on nuScenes under the $N_{b6n4}$ object detection (OD) and \data attribute detection (AD) setting. Acronyms: PSP: perspective specified prompt; HFA: horizontally flip augmentation; FM: foundation model; SR: success rate.}
\label{tab:ablation_attribute_prompt}
\end{wraptable}
Tab.~\ref{tab:ablation_attribute_model} shows that OneLLM significantly outperforms the other alternatives across all metrics.
This represents substantial gain over CLIP and CogVLM.
Even in novel class detection ($\text{AP}_{\text{N20}}$), the performance gap is pronounced. We attribute OneLLM's superior performance over CLIP and CogVLM to the additional FM encoder for point cloud which is particularly suitable for the LiDAR input modality.

\noindent\textbf{Using clearer descriptions \& augmenting by horizontally flipped visual features.}
We evaluate the efficacy of perspective-specified prompts (PSP) and horizontally flip augmentation (HFA) via the ablation study in Tab.~\ref{tab:ablation_attribute_prompt}. 
Using CLIP as the FM, we find that:
(1) PSP alone improves mAP by 0.41\% and attribute detection SR by 0.63\%. 
This gain supports the benefit of providing more descriptive prompts that help the model better distinguish between objects based on perspective, enhancing its accuracy.
(2) HFA alone yields larger gains, with mAP improving by 0.53\% and attribute detection SR improving by 1.23\%. 
These results show that HFA makes \ours's detection more robust.
(3) Jointly using PSP and HFA achieves even better performance than using each individually.
Finally, integrating PSP and HFA with OneLLM instead of CLIP achieves the best performance across all metrics.
These results suggest that viewpoint prefixes in prompts could ensure spatial direction remains disambiguated. \ours with OneLLM, when augmented with PSP and HFA, provides a stronger baseline for the OV 3D object and attribute detection task, likely due to OneLLM’s richer, more contextually aware embeddings.

\section{Conclusion}
\label{sec:conclusion}

We propose \ours, a novel framework enabling open-vocabulary (OV) multimodal 3D object detection with attribute detection, requiring no novel class information. 
It uses foundation model features and prompt tuning to bridge 3D features and text descriptions, while jointly detecting attributes like spatial relationships and motion states. 
We introduce \data with comprehensive attribute annotations for evaluating OV attribute detection. 
On nuScenes and Argoverse 2, \ours outperforms SOTA OV 3D object detection methods while successfully detecting object attributes.

\bibliography{egbib}

\clearpage
\setcounter{page}{1}
\setcounter{section}{0}
\setcounter{table}{8}
\setcounter{figure}{5}
\maketitlesupplementary

\section{Vocabulary Settings}
\label{sec:setting}

Following the rule of open-vocabulary setting~\cite{survey, CoDAv2, Find_n_Propagate}, we designed the vocabulary sets for the object and attribute detection, which are used during training.  
Following prior work~\cite{CoDAv2, OpenScene}, the vocabulary sets contain all existing base and novel classes, as well as additional classes.
The size of the vocabulary set determines the size of the text feature during the text-visual feature alignment, which further decides the size of spawned semantic space.
At testing time, the vocabulary set is replaced only by the union of all the existing base and novel classes.
Tab.~\ref{tab:setting} and Tab.~\ref{tab:setting_a} show the object and attribute vocabulary sets we used, respectively.

\begin{table}[ht]
\centering
\begin{minipage}[t]{0.49\linewidth}
\centering
\resizebox{\linewidth}{!}{
\begin{tabular}{cc}
\toprule
dataset & vocabulary set for object detection\\
\midrule
nuScenes  & \makecell{Car, Construction vehicles, Trailer, Barrier, Bicycle\\ Pedestrian, Truck, Bus, Motorcycle, Traffic cone \\
Animal, Ambulance, Police, Pushable pullable object \\
Debris, Bicycle rack} \\
\midrule
Argoverse 2 & \makecell{Regular Vehicle, Trailer, Bicycle, Pedestrian \\ 
Truck, Bus, Motorcycle, Construction cone \\
Animal, Bollard, Sign, Large vehicle \\
Wheeled device, Stroller, Railed vehicle}\\
\bottomrule
\end{tabular}  
}
\caption{The vocabulary sets we used for open-vocabulary object detection on different datasets. 
Following the rule of open-vocabulary setting~\cite{survey, CoDAv2, Find_n_Propagate}, our object vocabulary set contains all existing classes in base and novel objects as well as all rest object classes defined in nuScenes and Argoverse 2 dataset.}
\label{tab:setting}
\end{minipage}
\hfill
\begin{minipage}[t]{0.49\linewidth}
\centering
\vspace{-1.5cm}
\resizebox{\linewidth}{!}{
\begin{tabular}{cc}
\toprule
dataset & vocabulary set for attribute detection\\
\midrule
\textit{\data}  & \makecell{\textcolor{teal}{with rider}, \textcolor{teal}{without rider}, \textcolor{violet}{moving}, \textcolor{violet}{standing}, \\ \textcolor{violet}{sitting lying down}, \textcolor{brown}{parked}, 
\textcolor{brown}{moving}, \textcolor{brown}{stopped}, \\ 
in front of, behind, on the left of, on the right of} \\
\bottomrule
\end{tabular}
}
\caption{The vocabulary set we used for open-vocabulary attribute detection on our \data dataset. 
The attributes colored with \textcolor{teal}{teal}/\textcolor{violet}{violet}/\textcolor{brown}{brown} are the attributes exclusively associated with \textcolor{teal}{cycle}/\textcolor{violet}{pedestrian}/\textcolor{brown}{vehicle} classes.
Following the rule of open-vocabulary setting~\cite{survey, CoDAv2, Find_n_Propagate}, our attribute vocabulary set contains all existing classes in base and novel attribute as well as all the rest of attribute classes defined in the \data dataset.  }
\label{tab:setting_a}
\end{minipage}
\end{table}

\newpage
\section{More Details on Dataset \& Metrics}
\label{sec:dataset_eval_supp}

In nuScenes~\cite{nuScenes}, we follow the official split of nuScenes, which contains 1000 driving scenes captured in complex urban environments, divided into 700 for training, 150 for validation, and 150 for testing. 
In Argoverse 2~\cite{Argoverse2}, we use a similar train/val/test split (700/150/150) as nuScenes. 
As detailed in Section~\ref{sec:setting} of the supplementary, our vocabulary set includes all base and novel classes plus additional dataset-defined labels, following the OV setting protocols from~\cite{survey, CoDAv2, Find_n_Propagate}. During training, the full vocabulary is used for text embedding; during testing, only base and novel classes are retained.
\\

\noindent In the task of Object Detection, we report:
\begin{itemize}[leftmargin=1.5em]
    \item \textbf{mAP}: mean Average Precision with standard 3D IoU thresholds (0.2 for Argoverse 2, nuScenes follows official setup).
    \item \textbf{NDS}: nuScenes Detection Score, combining mAP with additional metrics such as translation, scale, orientation, velocity, and attribute accuracy.
    \item \textbf{$\text{AP}_\text{N}$}: mean AP computed only over novel classes to evaluate generalization under OV settings.
\end{itemize}

\noindent In the task of Attribute Detection, we use \textbf{Success Rate (SR)}:
\begin{itemize}[leftmargin=1.5em]
    \item \textbf{SR (AD only)}: measures the percentage of correctly classified attribute labels among all attribute-annotated proposals.
    \item \textbf{SR (AD \& OD)}: measures success rate conditioned on correct object category and localization.
\end{itemize}
The classification threshold is set to 0.5 for all attribute categories unless otherwise specified.

\section{Implementation details}
\label{sec:imp}
Implementing \ours in PyTorch~\cite{PyTorch}, we use AdamW optimizer with $\beta_1$=0.9, $\beta_2$=0.95, and weight decay of 0.1.
We set the number of object queries to 128 for both nuScenes and Argoverse 2. 
Initially, we train a base 3DETR model for 20 epochs using only class-agnostic distillation. 
Then, the model continues to be trained for 20 epochs.
The hyper-parameters used during training follow the default 3DETR configuration specified in~\cite{3detr, CoDAv2}.

\newpage
\section{More Qualitative Results}
\label{sec:vis}

\begin{figure}[ht]
    \centering
    \includegraphics[width=.995\linewidth]{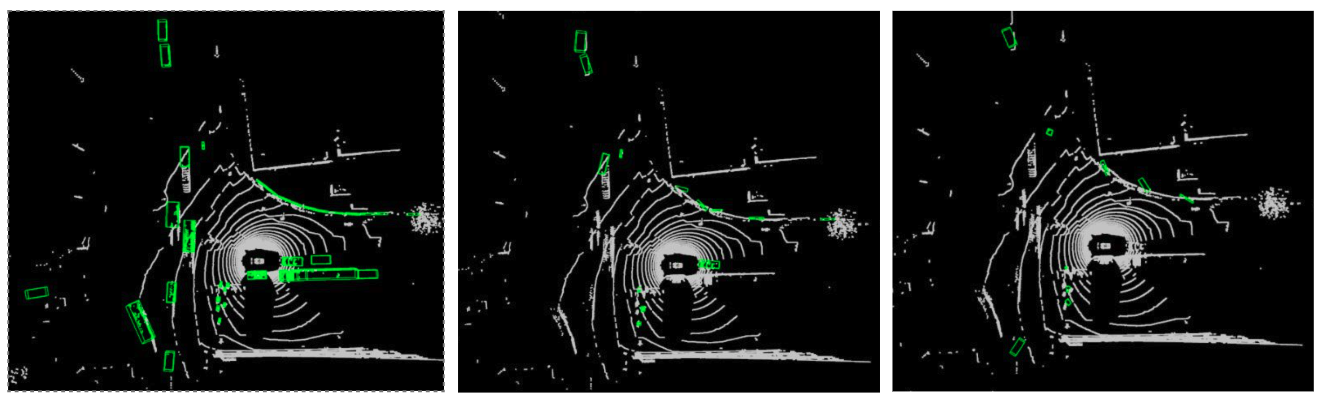}
    \caption{Qualitative comparison of \ours (middle) versus CoDAv2~\cite{CoDAv2} (right) with the ground truth (left). 
    }
    \label{fig:qualitative0}
    \vspace{-.3cm}
\end{figure}

\begin{figure}[ht]
    \centering
    \includegraphics[width=.95\linewidth]{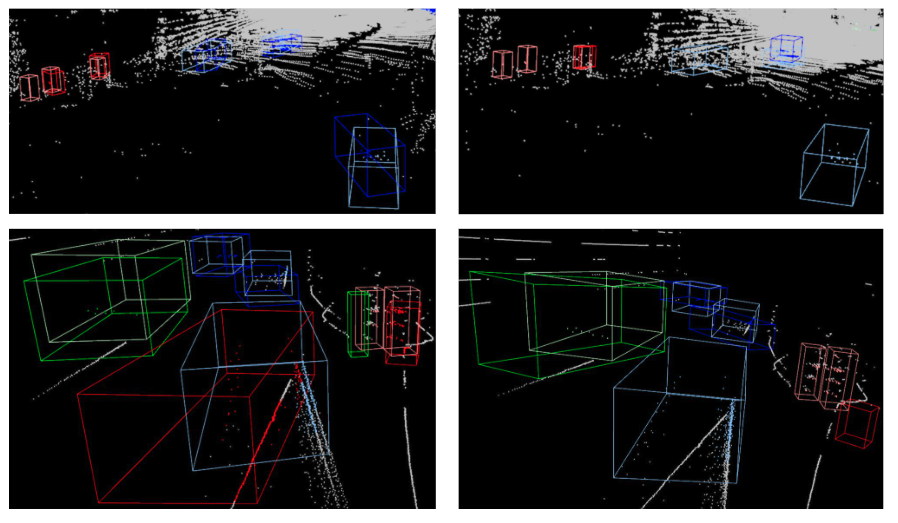}
    \caption{The qualitative comparison of \ours (left) versus CoDAv2~\cite{CoDAv2} (right) for 3D single object detection in nuScenes dataset. The ground truth annotations are rendered in \textcolor{cyan}{light blue}/\textcolor{pink}{light red}/\textcolor{green}{light green} for the class \textcolor{blue}{car}/\textcolor{red}{pedestrian}/\textcolor{teal}{others}, the predicted bounding boxes are rendered in \textcolor{blue}{blue}/\textcolor{red}{red}/\textcolor{teal}{green} for the class \textcolor{blue}{car}/\textcolor{red}{pedestrian}/\textcolor{teal}{others}.
    }
    \label{fig:qualitative}
    \vspace{-.3cm}
\end{figure}

Fig. \ref{fig:qualitative0} and Fig.~\ref{fig:qualitative} show the class-agnostic and class-specific qualitative comparison between \ours and CoDAv2, respectively.
Both qualitative results show that \ours's prediction is closer to the ground truth. 

\begin{figure*}[ht]
    \centering
    \includegraphics[width=1.0\linewidth]{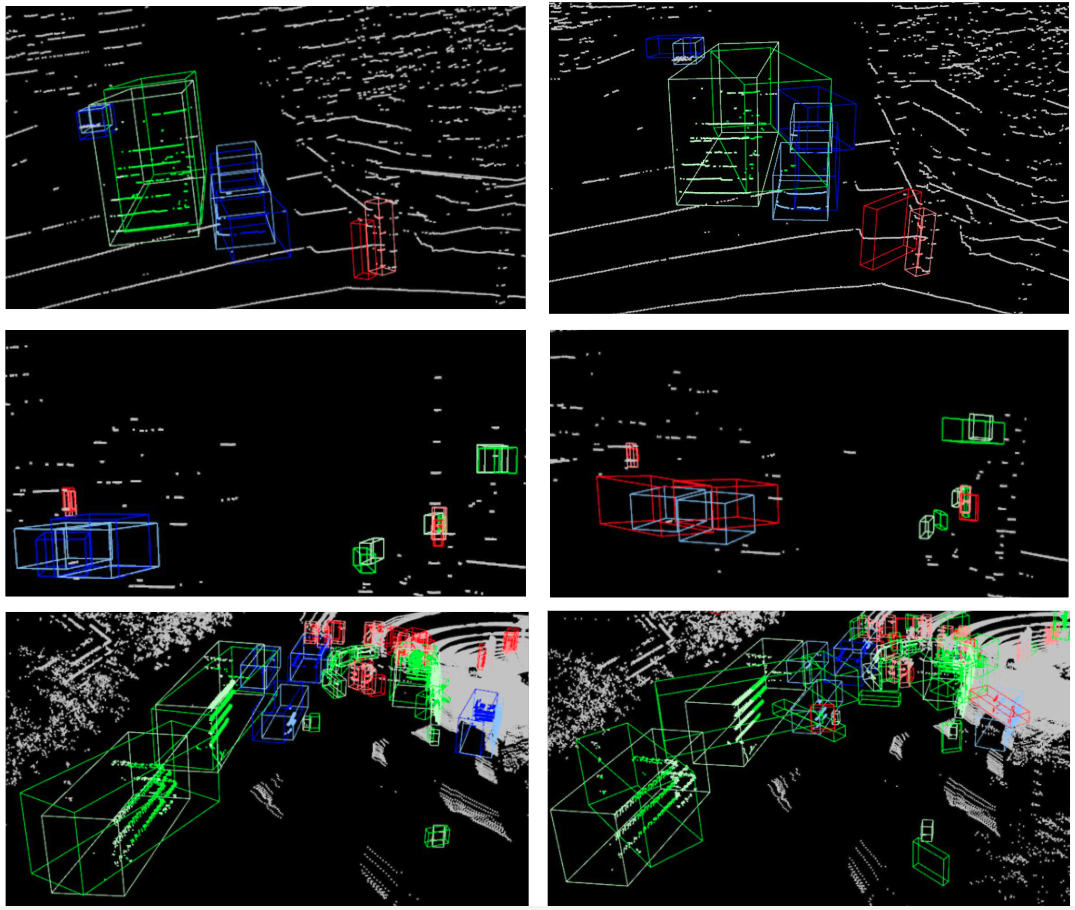}
    \caption{More qualitative comparison of \ours (left) versus CoDAv2~\cite{CoDAv2} (right) for 3D single object detection in nuScenes dataset. The ground truth annotations are rendered in \textcolor{cyan}{light blue}/\textcolor{pink}{light red}/\textcolor{green}{light green} for the class \textcolor{blue}{car}/\textcolor{red}{pedestrian}/\textcolor{teal}{others}, the predicted bounding boxes are rendered in \textcolor{blue}{blue}/\textcolor{red}{red}/\textcolor{teal}{green} for the class \textcolor{blue}{car}/\textcolor{red}{pedestrian}/\textcolor{teal}{others}.
    }
    \label{fig:qualitative-a1}
    \vspace{-.3cm}
\end{figure*}

\begin{figure*}[ht]
    \centering
    \includegraphics[width=1.0\linewidth]{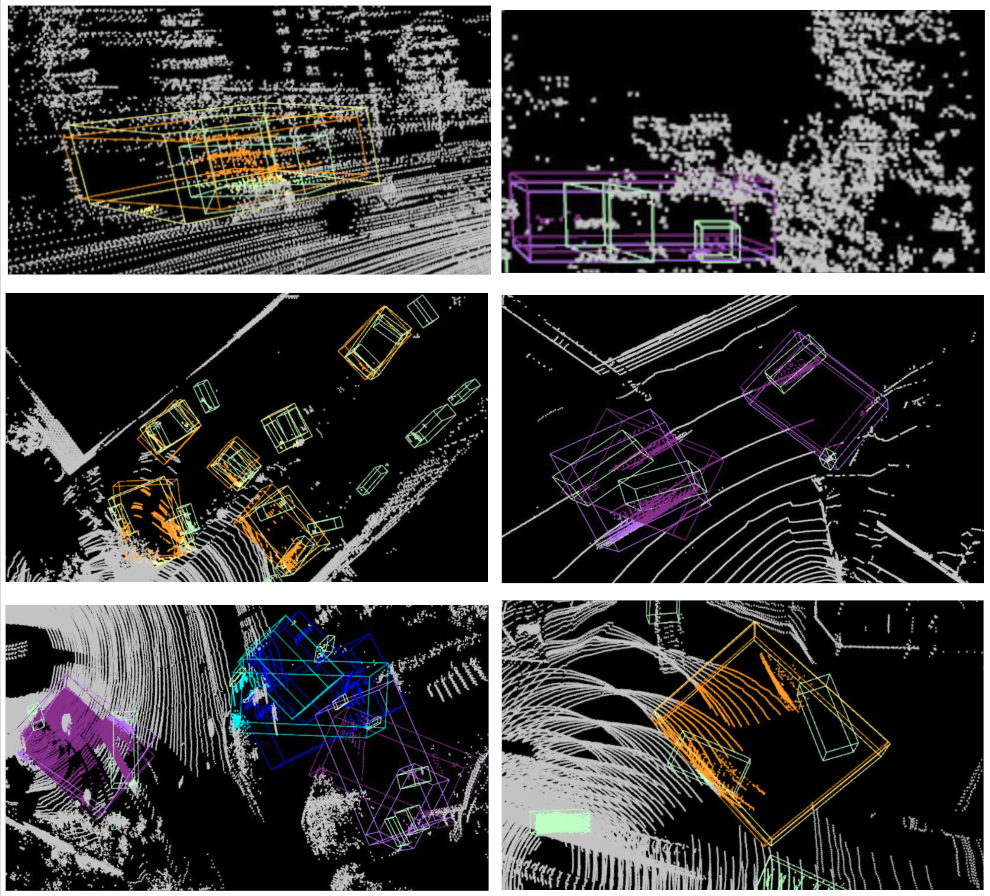}
    \caption{More qualitative results of \ours for 3D complex event detection in nuScenes dataset. All ground truth annotations of single object are rendered in \textcolor{green}{light green}. The ground truth annotations are rendered in \textcolor{pink}{light purple}/\textcolor{yellow}{yellow}/\textcolor{cyan}{light blue} for the 
    \textcolor{violet}{car-car}/\textcolor{orange}{pedestrain-pedestrain}/\textcolor{blue}{others} complex events, the predicted bounding boxes are rendered in \textcolor{violet}{purple}/\textcolor{orange}{orange}/\textcolor{blue}{blue} for the \textcolor{violet}{car-car}/\textcolor{orange}{pedestrain-pedestrain}/\textcolor{blue}{others} complex events. Examples of \textcolor{violet}{car-car}/\textcolor{orange}{pedestrain-pedestrain}/\textcolor{blue}{others} complex events can be: \textcolor{violet}{a car in front of the car}/\textcolor{orange}{a pedestrain on the left of the pedestrain}/\textcolor{blue}{a cyclist behind a car}.
    }
    \label{fig:qualitative-a2}
    \vspace{-.3cm}
\end{figure*}

We show more qualitative comparisons between \ourso (left) versus CoDAv2~\cite{CoDAv2} (right) for 3D single object detection on the nuScenes dataset in Fig.~\ref{fig:qualitative-a1}, where \ours’s prediction is closer to the ground truth compared with CoDAv2.
We show more qualitative results of \ourso for 3D complex event detection on the nuScenes dataset in Fig.~\ref{fig:qualitative-a2}, where \ours can successfully detect complex events.

\newpage

\newpage


\end{document}